\documentclass{article}

\usepackage{PRIMEarxiv}

\usepackage[utf8]{inputenc} 
\usepackage[T1]{fontenc}    
\usepackage{hyperref}       
\usepackage{url}            
\usepackage{booktabs}       
\usepackage{amsfonts}       
\usepackage{nicefrac}       
\usepackage{microtype}      
\usepackage{lipsum}
\usepackage{fancyhdr}       
\usepackage{graphicx}       
\usepackage{multirow}

\graphicspath{{media/}}     

\title{Enhancing Machine Learning Model Performance with Hyper Parameter Optimization
\thanks{\textit{\underline{Citation}}: 
\textbf{DOI:}} 
}

\author{
  Caner Erden \\
  AI Research and Application Center,  \\
  Sakarya University of Applied Sciences \\
  Sakarya, Türkiye\\
  \texttt{cerden@subu.edu.tr} \\
   \And
  Halil İbrahim Demir, Abdullah Hulusi Kökçam \\
  Department of Industrial Engineering \\
  Sakarya, Sakarya University \\
  Türkiye\\
  \texttt{\{hidemir, akokcam\}@sakarya.edu.tr} \\
}

\begin{document}
\maketitle

\begin{abstract}
One of the most critical issues in machine learning is the selection of appropriate hyper parameters for training models. Machine learning models may be able to reach the best training performance and may increase the ability to generalize using hyper parameter optimization (HPO) techniques. HPO is a popular topic that artificial intelligence studies have focused on recently and has attracted increasing interest. While the traditional methods developed for HPO include exhaustive search, grid search, random search, and Bayesian optimization; meta-heuristic algorithms are also employed as more advanced methods. Meta-heuristic algorithms search for the solution space where the solutions converge to the best combination to solve a specific problem. These algorithms test various scenarios and evaluate the results to select the best-performing combinations. In this study, classical methods, such as grid, random search and Bayesian optimization, and population-based algorithms, such as genetic algorithms and particle swarm optimization, are discussed in terms of the HPO. The use of related search algorithms is explained together with Python programming codes developed on packages such as Scikit-learn, Sklearn Genetic, and Optuna. The performance of the search algorithms is compared on a sample data set, and according to the results, the particle swarm optimization algorithm has outperformed the other algorithms.
\end{abstract}

\keywords{Meta-heuristics \and hyper parameter optimization \and Bayesian optimization \and genetic algorithm \and particle swarm optimization}

\section{Introduction}
Thanks to the machine learning (ML),and deep learning (DL) methods in artificial intelligence studies, a wide range of applications and publications have been developed in many areas, such as healthcare \cite{ahmed_adaptive_2022}, finance \cite{rundo_machine_2019}, and transportation \cite{bhavsar_machine_2017}. Organizations using ML come up with problems by making data-based interpretations and analyses. Although ML methods are easily implementable, the development of successful, consistent, and generalized models is a complex process. There are several alternatives for increasing the learning performance in machine-learning problems. Which are; (i) Using more data and increasing the amount of training data allows the model to generalize more effectively. (ii) Data preprocessing: Data preprossessing methods such as normalizing data, selecting features, or scaling features can improve the performance of the model. (iii) Model selection: Choosing the best appropriate model among different ML algorithms can increase the learning performance, as well. For example, a researcher might try an artificial neural network (ANN) model instead of a decision tree model. (iv) Ensemble methods: The learning performance can be increased by using multiple ML models or by combining the results of trained models. (v) Hyper parameter optimization: Adjusting the hyper parameters to optimal values can increase the learning performance. For example, the learning rate or initial values of the weights can be changed. This process aims to obtain more accurate models through various experimental studies. One of the challenging issues in ML is hyper parameter optimization(HPO). With ML models integrating with an optimization algorithm, models with acceptable performance that are close to the best are being developed. The main objectives of the HPO are as follows:

\begin{itemize}
\item To increase the generalization ability of models,
\item To achieve a higher train or test performance score,
\item To capture the best combination of parameters without trying to search space entirely,
\item To avoid over fitting to the training data as well as under fitting.
\end{itemize}

According to several studies, swarm intelligence and evolutionary computation can be used to increase the predictability and generalized of ML models. In this study, a comparison study will be conducted between meta-heuristic algorithms such as the genetic algorithm(GA), particle swarm optimization (PSO), and classical HPO techniques such as grid, random search, and Bayesian optimization.

\section{Understanding Hyper parameters}
\label{sec:2}

There are two types of parameters in ML: (i) model parameters and (ii) hyper parameters (HPs). Model parameters can generally vary depending on the variables in the data set and may not be configure by the user. The model performs the learning process by changing them during the training. For example, the weights in ANNs are model parameters. The second parameter type is HPs, which are set before training a ML model and cannot be directly learned from the data. These parameters control the learning process, such as the learning rate, maximum depth in a decision tree, and number of neurons in an ANN layer. The optimal values of these parameters can significantly affect the performance of the model, including its accuracy, speed, and generalization to new data. The main objective of the HPO is to effectively scan the search space to find the values that result in the best performance on a validation or test set. This process is typically performed through iterative steps that attempt different combinations of HPs and evaluate their performance, thereby allowing the selection of the best set of HPs for a given task.

Suppose the optimization process in the form of a set of all HPs $\sigma=x_1,x_2,\ldots,x_{nhp}$. $nhp$ denotes the number of hyper parameters. $x_i\in\ X_i$, $X_i$ is the search space of $x_i$. $X_i$ may be sometimes categorical (such as criterion in a random forest (RF)), integer (such as max\_depth), continuous (such as learning rate) variable. In the case of continuous or integer, the spaces can be selected from a probability distribution. HPs are best when $f\left(\sigma\right)$ is the minimum or maximum according to the performance metric. If the performance metric is the error values, the objective function is $min\ f\left(\sigma\right)$, otherwise it is $max\ f\left(\sigma\right)$ as shown in Eq. \ref{Eq1} \cite{feurer_hyperparameter_2019}. Another challenging in HPO is because of the interconnected hyper parameters. For example, in deep neural networks, the behavior of HPs such as learning rate will change according to the optimizer hyper parameter selected. At the same time, the learning rate parameter that is valid when the optimizer is Adam may not apply to any other optimizer. All these situations are among the difficulties of HPO studies. ML engineer will need to solve many complex challenges like those and offer practical solutions to these problems.
\begin{equation}
\label{Eq1}
\sigma^\ast={\rm argmin}_{x_i\in X_i}\ f\left(\sigma\right)\ or\ {\rm argmax}_{x_i\in X_i}\ f\left(\sigma\right)
\end{equation}

\begin{figure}
  \centering
  \includegraphics[width=10cm]{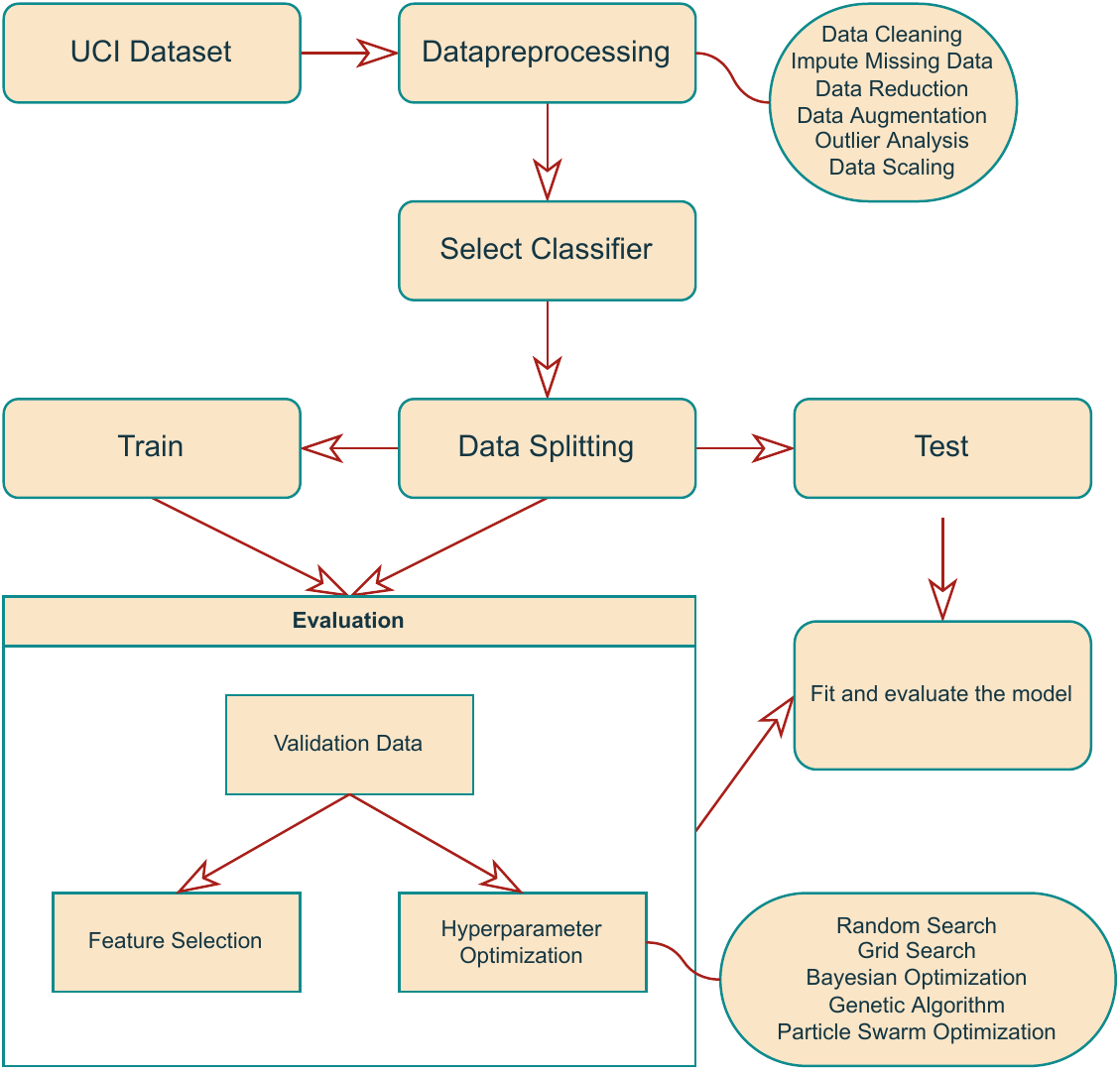}
  \caption{The training cycle of ML}
  \label{fig:fig1}
\end{figure}

HPO studies are used in all three types of machine learning: supervised, unsupervised, and reinforced. In supervised learning, the labels of output variables are known. As a result of the learning, performance criteria such as confusion matrix, precision, ROC-AUC curves, and accuracy were developed to test the accuracy of the training. In regression analysis, the metrics include the mean square error (MSE), mean absolute error (MAE), and R2. Two sets were selected as the training and test sets to evaluate their performance. Training was performed on the training set, and if the performance of the training provided the desired result, training was applied to the test set. In Fig. \ref{fig:fig1}, a flowchart of machine learning is presented. 

Although the flow of ML studies seems simple, it requires routine and challenging processes because it involves repetitive operations. Automated machine learning (AutoML) studies aim to produce machine learning models that are more suitable for real data with little intervention in the model \cite{bello_neural_2017}.  AutoML studies include sub-studies such as selecting the learning algorithm, determining the set of features, and HPO. Simultaneously, AutoML studies were conducted to obtain repeatable results.

\begin{figure}
  \centering
  \includegraphics{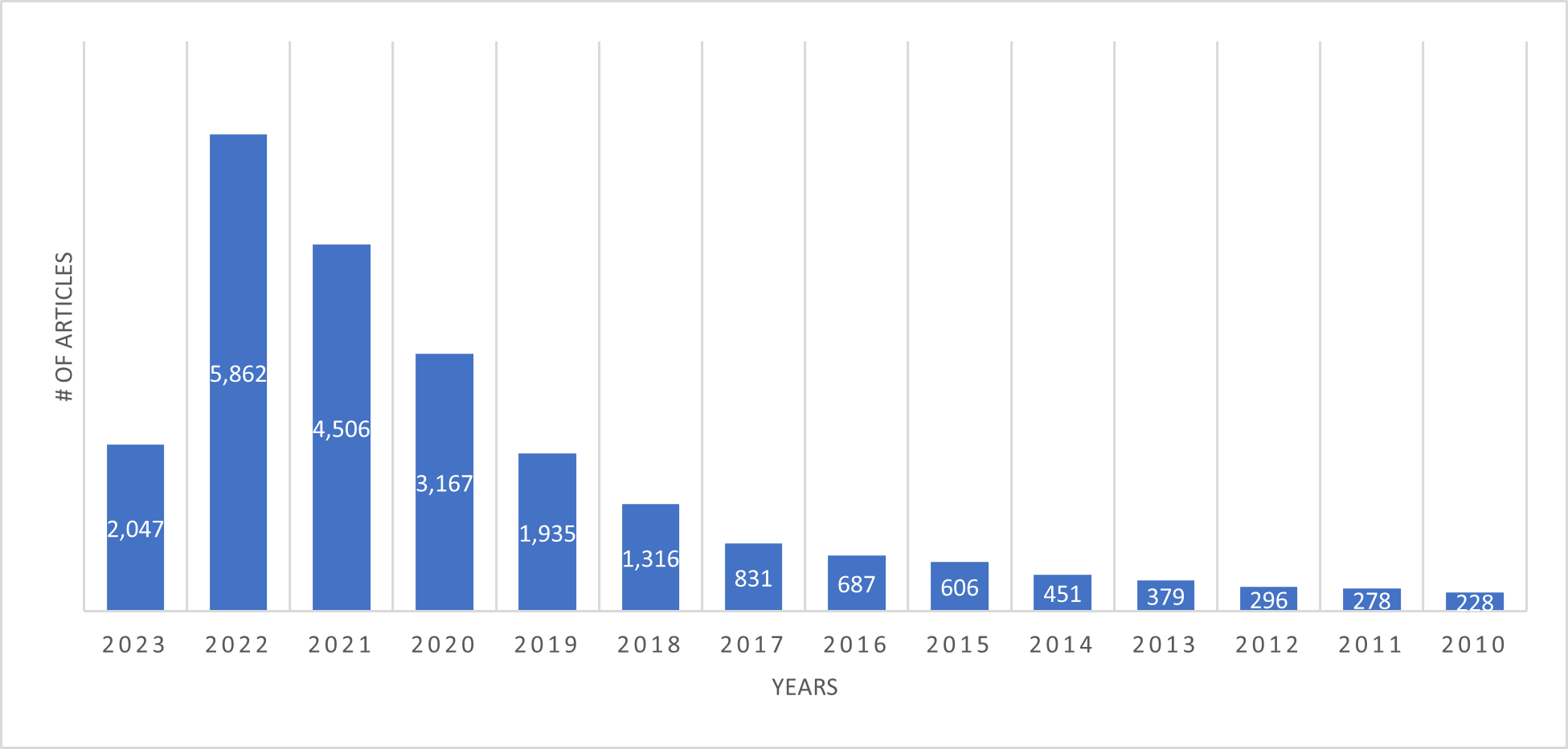}
  \caption{HPO Studies in the literature}
  \label{fig:fig2}
\end{figure}

In recent years, we have witnessed DL applications in many areas such as image recognition, video processing, voice recognition, and self-driving cars, which have not been realized in the past. Tuning the HPs of deep-learning algorithms is a very difficult process for users. Therefore, HPO is indispensable in DL studies. As can be expected, the topic of HPO has gained significant momentum with the increase in DL studies owing to the more complex structure of DL. In this context, when the science direct database is scanned with the "hyper parameter optimization" query, the number of articles can be seen in Fig. \ref{fig:fig2}. This is a topic of growing interest.

The research questions usually focus on the following: (i) Is it possible for the DL model to achieve a better performance with hyper parameter changes? (ii) What are the best hyper parameter combinations for the DL model? To answer these questions, it is necessary to use several optimization techniques and correctly define the constraints together with the objective function.

\begin{figure}
  \centering
  \includegraphics[width=0.9\textwidth]{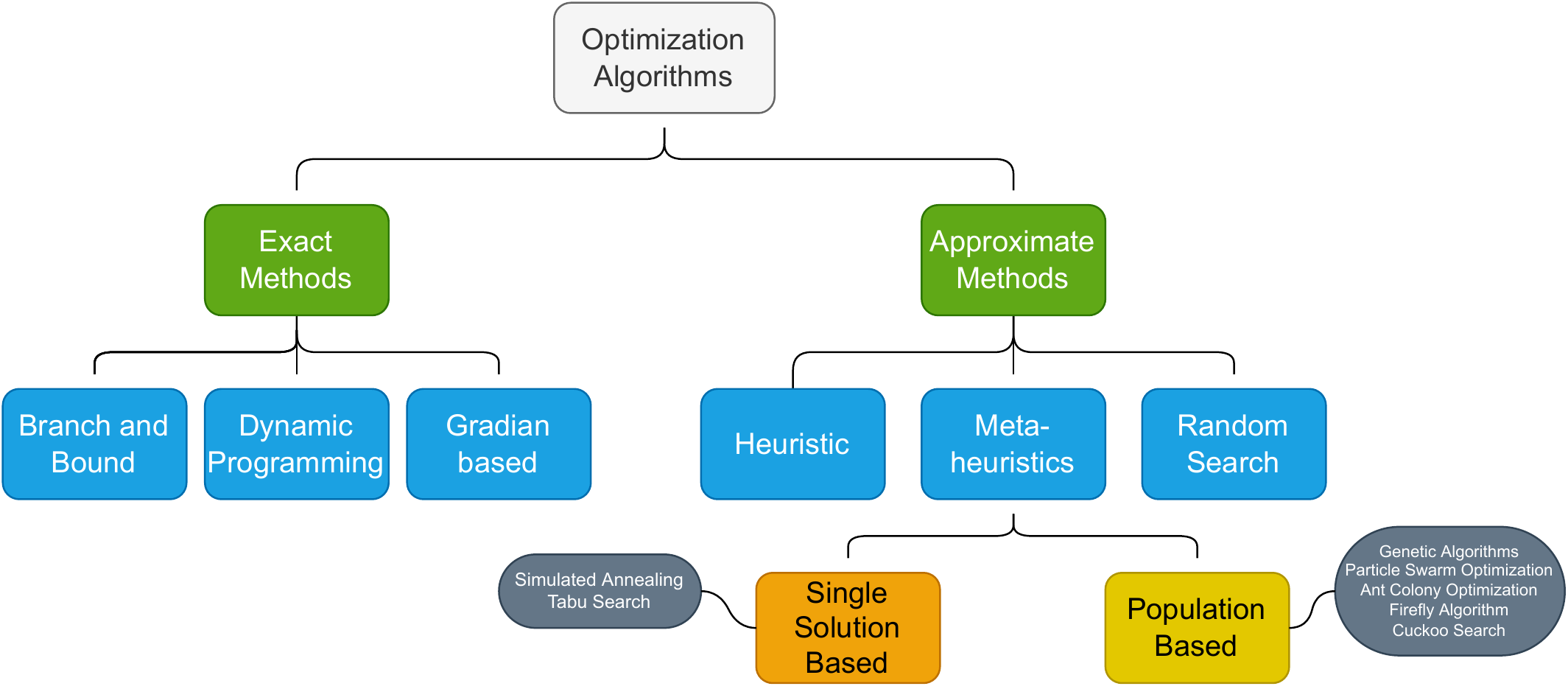}
  \caption{The training cycle of ML}
  \label{fig:fig3}
\end{figure}

To classify the optimization techniques used for HPO, a taxonomy can be determined in general terms, as shown in Fig. \ref{fig:fig3}. According to this classification, there are a variety of optimizations that provide the exact best results and approximate good results. It is impossible to use algorithms that provide precise results in HPO studies. Instead, approximate methods that work well are classified as local and meta-heuristic algorithms. Meta-heuristic algorithms, on the other hand, are classified as evolutionary algorithms (such as genetic algorithms (GAs), harmonic search, genetic programming), population-based algorithms (such as ant colony optimization (ACO), artificial bee colony (ABC), PSO), physics-based (such as electromagnetism-based, annealing simulation, gravitational search), and human-based algorithms.

\section{Hyper parameter Optimization Techniques}

In HPO, most studies have been conducted using trial and error, grid search, and random search methods. The trial-and-error method requires primitive and challenging tasks. The researcher’s skills are at the forefront of this technique.

Trial and Error Technique: The model was trained with different HPs and its performance was tested. In conclusion, HPs that provide the best performance are used.

\textbf{Grid search}, is a simple approach for HPO. This involves exhaustively searching the hyper parameter space by testing all possible combinations of hyper parameters. An ML model was trained and evaluated for each combination, and its performance was recorded. The combination of HPs that resulted in the best performance on the validation set was selected as the optimal set. The grid search is simple to implement, easy to understand, and can be used as a baseline comparison to other HPO techniques. However, grid search can be computationally expensive, especially for high-dimensional hyper parameter spaces, and may not always find the optimal set of hyper parameters, particularly when the relationship between HPs and performance is not well understood. Despite its limitations, grid search remains a popular method for hyper parameter optimization, particularly for small data sets and simple models.

\textbf{Random Search} is another popular approach to HPO in machine learning. Unlike grid search, which exhaustively searches the hyper parameter space, random search generates a set of random combinations of HPs and evaluates their performance. The goal is to randomly sample the hyper parameter space and determine the best set of hyper parameters. Random search is computationally less expensive than grid search and can be more efficient for high-dimensional hyper parameter spaces. Additionally, random search has been shown to perform better than grid search in some cases, particularly when the relationship between HPs and performance is not well understood. However, the performance of random search is highly dependent on the number of trials and random sampling distribution, and it may not always find the optimal set of hyper parameters. Despite these limitations, random search remains a popular method for hyper parameter optimization, particularly when combined with Bayesian optimization or other meta-heuristic algorithms.

\textbf{Bayesian optimization} is a probabilistic model-based approach to HPO in machine learning. It models the relationship between HPs and performance using a Bayesian model such as a Gaussian process. This model was used to guide the search for an optimal set of hyper parameters. Bayesian optimization iterative selects the next set of HPs to be evaluated based on the model’s performance prediction, allowing for an efficient search of the hyper parameter space. Bayesian optimization is highly effective for hyper parameter optimization, particularly in high-dimensional spaces, and it is widely used in machine learning and other fields. However, Bayesian optimization can be more computationally expensive than grid search or random search and may require more computational resources for implementation. Despite these limitations, Bayesian optimization remains a popular method for hyper parameter optimization, particularly when combined with other optimization techniques or meta-heuristics.

The other category of algorithms in the topic of HPO is meta-heuristic algorithms. The fact that meta-heuristics do not require gradient knowledge and may be used on issues where the gradient cannot be collected is one of their key benefits in optimization problems. They may also be used to solve optimization issues in which the fitness or objective function is unknown or differentiable, which is an extra benefit. In addition, compared to other optimization methods, meta-heuristics can deliver an optimum solution using fewer processing resources \cite{kouziokas_swarm_2022}. Meta-heuristics can be used in any optimization problem as well as in some special areas, such as financial trading, credit card data, portfolio optimization, routing problems, load balancing problems, facility location, influence maximization in social networks, scheduling, smart building, and reconstruction of ECG signals \cite{jafar_financial_2022} \cite{shahriari_taking_2015}. Some studies in the field of HPO by meta-heuristics are listed in Table \ref{tab:tab1}. 

\begin{table}[hbt!]
\centering
\caption{Reference studies}
\label{tab:tab1}
\resizebox{\columnwidth}{!}{%
\begin{tabular}{|p{0.10\linewidth}| p{0.30\linewidth} | p{0.30\linewidth} |  p{0.2\linewidth} | p{0.10\linewidth} |}
\hline
\textbf{Reference} &
  \textbf{Optimization Algorithm} &
  \textbf{ML Algorithm} &
  \textbf{Application Area} &
  \textbf{ML/DL} \\ \hline
\cite{ahmet_senel_hyperparameter_2023} &
  Grey Wolf Optimization (GWO) &
  Deep learning models (CNN) &
  Galaxy Classification &
  DL \\ \hline
\cite{ahmed_adaptive_2022} &
  Simulated Annealing (SA) &
  extreme gradient boosting, categorical boosting &
  E-Triage Tool &
  ML \\ \hline
\cite{ali_hyperparameter_2023} &
  Ant Bee Colony Algorithm, GA, Whale Optimization, and   PSO &
  Support Vector Machine (SVM) &
  Optimizing the Computational Complexity &
  ML \\ \hline
\cite{erden_genetic_2023} &
  GA &
  Deep learning models (RNN, LSTM, GRU) &
  PM2.5 time-series prediction &
  DL \\ \hline
\cite{he_assessment_2023} &
  GWO, the whale optimization algorithm, swarm algorithm &
  RF &
  Predict Blast-Induced Over break &
  ML \\ \hline
\cite{inik_cnn_2023} &
  PSO &
  Deep learning models (CNN) &
  Environmental Sound Classification &
  DL \\ \hline
\cite{kalita_novel_2023} &
  Moth-Flame Optimization &
  SVM &
  Intrusion Detection System &
  ML \\ \hline
\cite{si_metamodel-based_2023} &
  non dominated sorting GA II, GA, PSO, simulated   annealing, and the multi-objective GA &
  ANN &
  Energy Optimization &
  ML \\ \hline
\cite{tayebi_performance_2022} &
  GA, differential evolution, ABC, GWO, PSO, and teaching   learning-based optimization &
  AdaBoost, RF, logistic regression, SVM, k-nearest   neighbors, and decision tree &
  Fraud Transactions Detection &
  ML \\ \hline
\cite{nematzadeh_tuning_2022} &
  GWO and GA &
  Averaged Perceptron, Fast Tree, Fast Forest, Light   Gradient Boost Machine, Limited memory Broyden Fletcher Goldfarb Shanno   algorithm Maximum Entropy, Linear SVM, and DL &
  Biological, Biomedical Data sets &
  ML and DL \\ \hline
\cite{ma_comprehensive_2022} &
  multi-verse optimization &
  SVM &
  Geohazard Modeling &
  ML \\ \hline
\cite{bacanin_benefits_2023} &
  GA, PSO, ABC, Firefly Algorithm, Bat Algorithm, Sine   Cosine Algorithm &
  Deep learning models (LSTM) &
  Energy Load Forecasting &
  DL \\ \hline
\cite{ma_metaheuristic-based_2022} &
  PSO &
  SVM &
  Landslide Displacement Prediction &
  ML \\ \hline
\cite{chou_metaheuristics-optimized_2022} &
  Jellyfish search optimization &
  Deep learning models (CNN) &
  Generation Of Sustainable Energy &
  DL \\ \hline
\cite{suddle_metaheuristics_2022} &
  GA, PSO, Differential Evolution, Firefly, and Cat Swarm   Optimization &
  Deep learning models (LSTM) &
  Sentiment Analysis &
  DL \\ \hline
\cite{drewil_air_2022} &
  GA &
  Deep learning models (LSTM) &
  Air Pollution Prediction &
  DL \\ \hline
\cite{al_duhayyim_metaheuristics_2022} &
  Salp Swarm Algorithm &
  Deep learning models (GRU) &
  Image Captioning System &
  DL \\ \hline
\cite{maumela_population_2022} &
  Ulimisana Optimization Algorithm &
  ML &
  ML Unfairness &
  ML \\ \hline
\cite{zhao_lithium-ion_2023} &
  Hunger Game Search &
  Gaussian process regression &
  Lithium-Ion Battery State of Health Estimation &
  ML \\ \hline
\cite{albakri_metaheuristics_2023} &
  Rock Hyrax Swarm Optimization &
  Deep learning models &
  Cybersecurity and Android Malware Detection and   Classification &
  DL \\ \hline
\end{tabular}
}
\end{table}

In addition, optimization algorithms can be used as optimizer for ML models. Examples include:

\begin{itemize}
\item \textbf{Stochastic Gradient Descent (SGD)} is a well-liked optimization technique in DL and ML. Finding the set of parameters that minimizes a loss function is the objective of optimization in these domains. SGD updates the parameters in the direction of the loss's negative gradient with respect to the parameters. Instead of computing the gradient using the whole training set as in batch gradient descent, the gradient is calculated using a single randomly chosen sample from the training set. The algorithm is stochastic since just one sample was chosen at random.
\item \textbf{Levenberg-Marquardt}: For non-linear least squares problems, the Levenberg-Marquardt algorithm is a potent optimization technique that has found widespread application in many disciplines, including machine learning, computer vision, and control engineering.
\item \textbf{Adagrad}: The stochastic gradient descent optimization algorithm Adagrad (Adaptive Gradient Algorithm) is applied to optimization issues in DL and machine learning. Adagrad's primary principle is to scale the learning rate for each parameter individually, so that parameters that receive a lot of updates have a reduced learning rate and parameters that receive less updates have a bigger learning rate. This makes it possible for the optimization process to continue moving forward for all parameters even when there are sparse gradients.
\item \textbf{RMSPop}: The optimization approach RMSProp (Root Mean Square Propagation) is used to train deep neural networks and other machine learning models. This stochastic gradient descent variation attempts to overcome some of the drawbacks of conventional gradient descent, including sluggish convergence and oscillating behavior.
\item \textbf{Adadelta}: It is a variant of the popular Adagrad algorithm and aims to address some of its limitations, such as the need to manually set the learning rate and the tendency to converge slowly.
\end{itemize}

\begin{figure}
  \centering
  \includegraphics{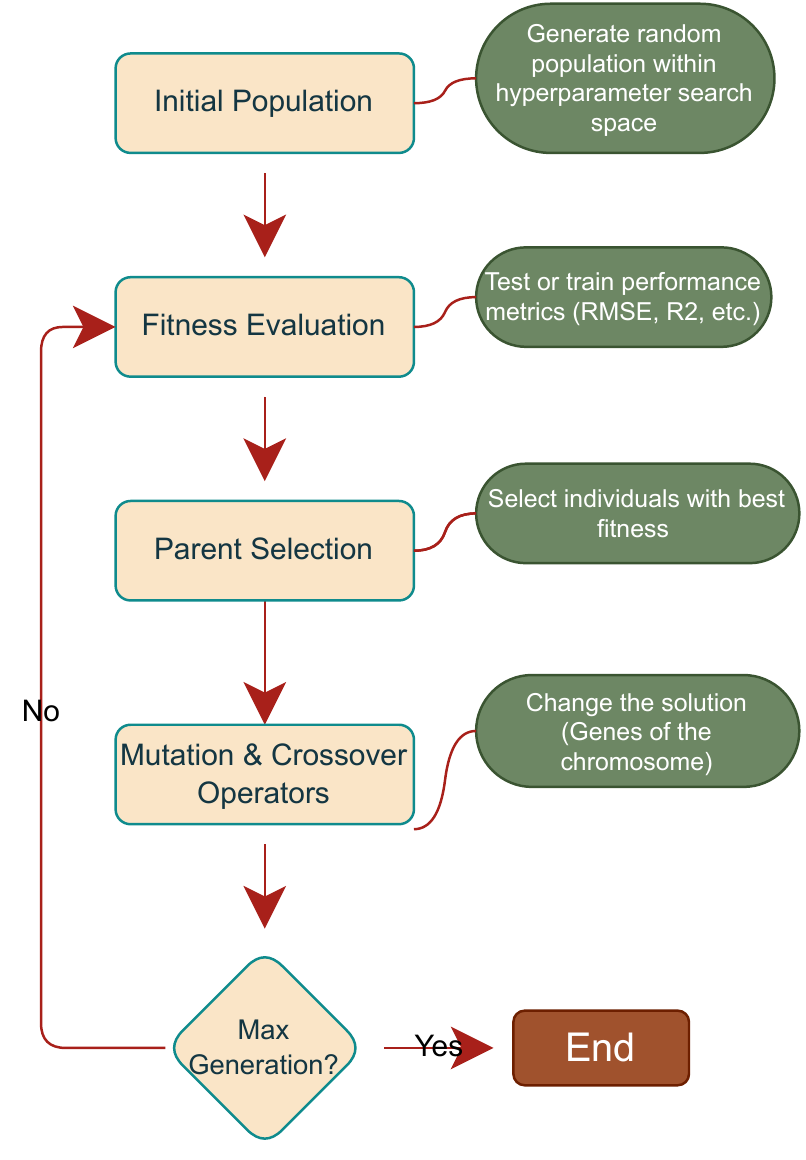}
  \caption{Genetic algorithm steps for hyper parameter optimization}
  \label{fig:fig4}
\end{figure}

\textbf{GAs}: GAs are meta-heuristic optimization methods inspired by the natural selection phase in biology. The GA works by encoding candidate solutions as binary strings or other representations and using operators such as mutation, crossover, and selection to generate new solutions and evolve the population over iterations. The performance of each candidate solution(fitness) was evaluated, and the best solutions were selected to form the next generation. This process continues until a satisfactory solution is found or a stopping criterion is met (see Fig. \ref{fig:fig4}).

\begin{figure}
  \centering
  \includegraphics{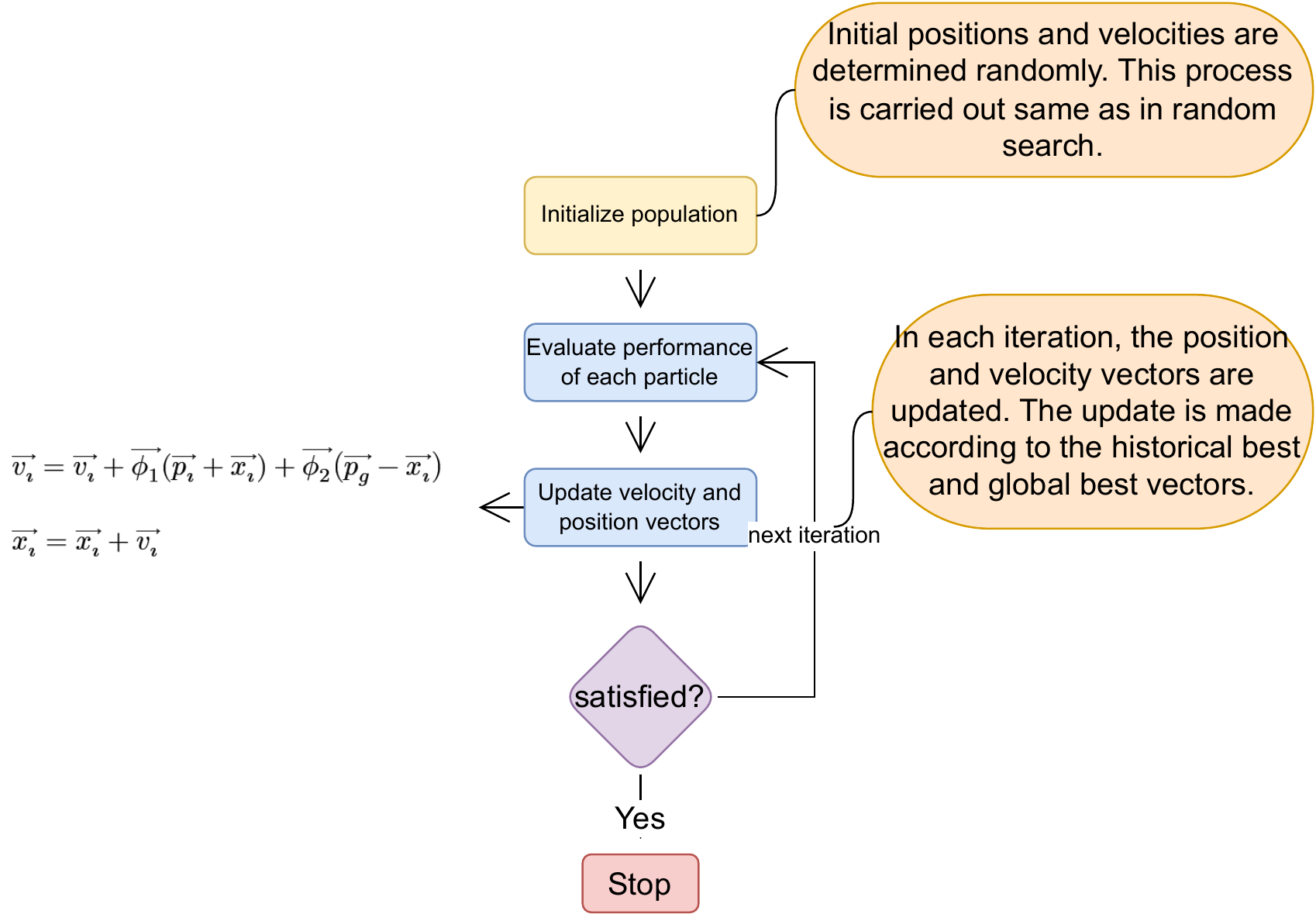}
  \caption{PSO for hyper parameter optimization}
  \label{fig:fig5}
\end{figure}

\textbf{PSO}: A meta-heuristic optimization technique, PSO, mimics the behavior of swarms such as fish and bird flocks. In PSO, a group of particles moves around the hyper parameter space, adjusting their locations according to their own and neighbors' best performance. The particles in the PSO algorithm are simulated in the search space adopted \cite{clerc_particle_2002}\cite{kennedy_particle_2010}. The flow of the PSO algorithm is shown in Fig. \ref{fig:fig5} \cite{noauthor_optunity_2023}. 

Where $v_i$ is the velocity, $x_i$ is the position, $p_i$ is the personal best position, $p_g$ is the global best position, $\phi_1$ and $\phi_2$ are the random values.

\textbf{Ant Colony Optimization}: ACO is a meta-heuristic optimization method inspired by the foraging behavior of ant colonies. It is often used for HPO in machine learning. ACO is computationally expensive and requires numerous evaluations to converge to an optimal solution. Studies of HPO with ACO are quite few \cite{lohvithee_ant_2021}. The studies to be carried out in this regard will be needed in the future.

\subsection{Available Tools for HPO}

To apply these techniques, a flexible tool is required. The most widely used programming language for machine learning applications is Python. Owing to its libraries, the Python programming language offers machine-learning applications more easily and flexibly. ML applications can also be developed on programs with interfaces, such as MATLAB, Rapidminer, and Weka. A comparison of popular ML tools is provided in Table \ref{tab:tab2}.

\begin{table}[hbt!]
\centering
\caption{Popular ML Tools}
\label{tab:tab2}
\resizebox{\columnwidth}{!}{%
\begin{tabular}{|p{0.1\linewidth}|p{0.2\linewidth}|p{0.20\linewidth}|p{0.1\linewidth}|p{0.40\linewidth}|}
\hline
 &
  \textbf{Platform} &
  \textbf{Fee} &
  \textbf{Programming Language} &
  \textbf{Usage} \\ \hline
Scikit-Learn &
  Linux,   Mac OS, Windows &
  Free &
  Python,   Cython, C, C++ &
  Classification,   Regression, Clustering, Data Preprocessing, Model Selection, Size reduction \\ \hline
PyTorch &
  Linux,   Mac OS, Windows &
  Free &
  Python,   C++, CUDA &
  Autograd   Module, Optimization Module, ANN Module \\ \hline
TensorFlow &
  Linux,   Mac OS, Windows &
  Free &
  Python,   C++, CUDA &
  Enables   data flow programming. \\ \hline
Weka &
  Linux,   Mac OS, Windows &
  Free &
  Java &
  Data   preparation Classification, Regression, Clustering, Visualization, Data   Mining, Relationship Rules \\ \hline
KNIME &
  Linux,   Mac OS, Windows &
  Free &
  Java &
  It   can work with large data volumes. It supports text mining and image mining   through plugins. \\ \hline
Colab &
  Cloud   Service &
  Free &
  - &
  PyTorch   supports Keras, TensorFlow, and OpenCV tools. \\ \hline
Apache Mahout &
  Cross-platform &
  Free &
  Java,   Scala &
  Preprocessing,   Regression, Clustering, Recommendation Systems, Distributed Linear Algebra,   Classification, regression, clustering, Hypothesis Testing \& Core   Methods, Image, Sound, and Signal. \& Processing \\ \hline
Shogun &
  Windows,   Linux, UNIX, Mac OS &
  Free &
  C++ &
  Regression,   Classification, Clustering, Support vector machines, Size reduction, Online   learning \\ \hline
Keras.io &
  Cross-platform &
  Free &
  Python &
  API   for ANN \\ \hline
Rapid Miner &
  Cross-platform &
  Yearly   Small: 2500$. Medium:5000 $. Large:10.000\$. &
  Java &
  Data   Loading and Conversion Data preprocessing and visualization. \\ \hline
\end{tabular}%
}
\end{table}

As can be seen from Table \ref{tab:tab2}, Python programming language stands out because of the library facilities it offers. One of most popular libraries in traditional machine learning applications is the Scikit-Learn library \cite{virtanen_scipy_2020}.  The Scikit-Learn library includes functions such as grid search and random search related to HPO. As previously mentioned, a random search includes combinations of random parameters. In the grid search, the result was obtained by trying all the parameter combinations. Many tools are available in the Python programming language with which HPO studies can be performed. Of the open-source tools mentioned here, except Auto-Weka, they are mainly developed in the Python programming language. These tools:

\begin{itemize}
\item Hyperopt (Hyperopt, 2011/2021) 
\item Auto-Weka (Automl/Autoweka, 2015/2021)
\item Auto-sklearn (Auto-Sklearn, 2015/2021)
\item Optuna (Optuna, 2018/2021)
\item Ray-Tune (Ray-Project/Ray, 2016/2021)
\item Bayesian Optimization (fernando, 2014/2021)
\end{itemize}

\section{Applications of Hyper Parameter Optimization}
\begin{figure}
  \centering
  \includegraphics[width=12cm]{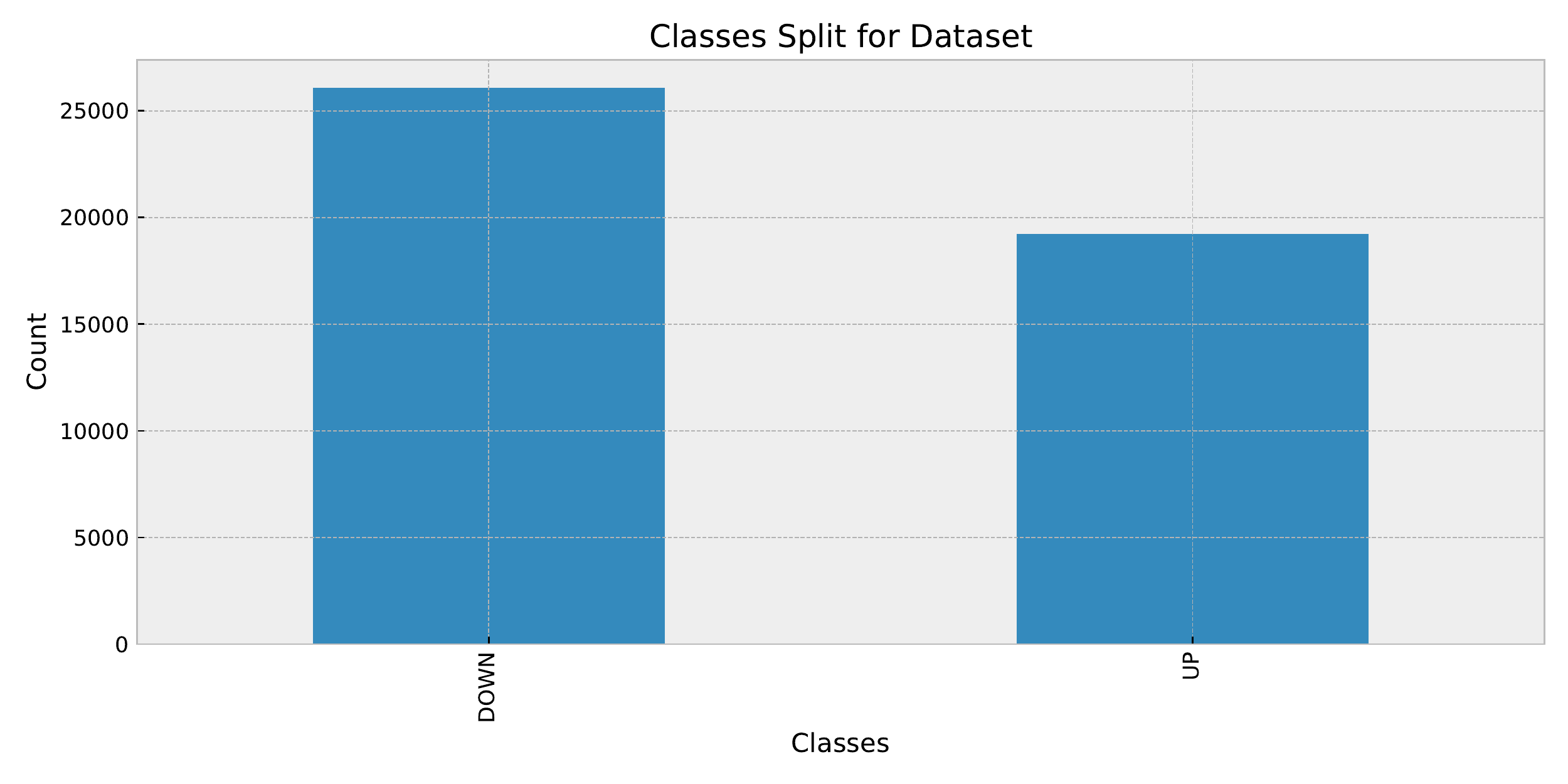}
  \caption{Number of instances for the output variable}
  \label{fig:fig6}
\end{figure}

The ELEC2 dataset is popular and can be accessed at https://datahub.io/machine-learning/electricity \cite{hutchison_learning_2004} \cite{harries_splice-2_1999}. For this study, a popular dataset was selected to make comparisons easier to interpret. The dataset contains 45312 data collected from the Australian New South Wales Electricity Market area from May 7, 1996, to December 5, 1998. Electricity prices vary according to the supply and demand. Each entry in the dataset had a duration of 30 min. The output variable of the dataset is whether the prices are up or down.  In this study, experiments are run on a data set that can be considered balanced. As seen in Fig. \ref{fig:fig6}, the total number of inputs with UP class is 19237 and the number of DOWN is 16075.

In this study, 20\% of the dataset was used for testing, and 80\% for training. MinMax normalization of variables in the dataset. In addition, training was carried out using 5-fold cross-validation. The performance of the algorithms was calculated using the accuracy score based on rates such as true positive(TP) true negative(TN), false positive (FP), false negative (FN), as presented in Eq. \ref{Eq2}. Where ${\hat{y}}_i$ is predicted, $y_i$ is the true value of the i-th sample, and $1\left(x\right)$ is the indicator function. Besides, for the performance evaluations of the algorithms, the precision, recall, and f1-score metrics were used as shown in Eq. 3-5.  Hyper parameter search spaces were determined to be fair for all algorithms. In this study, a RF algorithm was used. Therefore, in RF, the number of random features on each node (max\_features), the number of trees in the forest (n\_estimators), maximum depth (max\_depth), and the strategy the split will be performed (criterion) are used, and their search spaces are given in Table \ref{tab:tab3}.

\begin{equation}
\label{Eq2}
    \mathrm{accuracy}\left(y,\hat{y}\right)=\frac{1}{n_{\mathrm{samples}}}\sum_{i=0}^{n_{\mathrm{samples}}-1}1\left({\hat{y}}_i=y_i\right)
\end{equation}

\begin{equation}
    precision\ =\ \frac{TP}{TP+FP}\ 
\end{equation}

\begin{equation}
    recall\ =\ \frac{TP}{TP+FN}
\end{equation}

\begin{equation}
    f1\ Score\ =2\times\frac{Precision\ \times\ Recall}{Precision\ +\ Recall}
\end{equation}

\begin{table}[hbt!]
\centering
\caption{Hyper Parameter Search Spaces}
\label{tab:tab3}
\resizebox{\columnwidth}{!}{%
\begin{tabular}{|p{0.15\linewidth}|p{0.15\linewidth}|p{0.5\linewidth}|}
\hline
\textbf{HPs}           & \textbf{Grid Search}                  & \textbf{Random Search}, \textbf{Bayesian Optimization},   \textbf{Genetic Algorithm}, \textbf{PSO} \\ \hline
n\_estimators & {[}50, 100, 150, 200, 250{]} & Integer (50, 400)                                              \\ \hline
max\_features & {[}1, 3, 5, 8{]}             & Integer (4, 10)                                                \\ \hline
max\_depth    & {[}1, 2, 4, 8{]}             & Integer (4, 10)                                                \\ \hline
criterion     & {[}'entropy', 'gini'{]}      & {[}'entropy', 'gini'{]}                                        \\ \hline
\end{tabular}%
}
\end{table}

GA parameters:
\begin{itemize}
    \item Population Size: 10
    \item Generation Size: 40
    \item Crossover Probability: 0.9
    \item Mutation Probability: 0.05
    \item Algorithm: EaSimple
\end{itemize}
PSO parameters:
\begin{itemize}
	\item Population size: 10
	\item Generation Size: 40
	\item $\phi_1$, $\phi_2$: 0.5
\end{itemize}

The algorithms were implemented in the Python programming language using skopt \cite{head_scikit-optimizescikit-optimize_2021}, SciPy \cite{virtanen_scipy_2020}, sklearn (Auto-Sklearn, 2015/2021), sklearn\_genetic \cite{calzolari_manuel-calzolarisklearn-genetic_2022} and Optunity \cite{noauthor_optunity_2023} packages in the Jupyter notebook development environment on a laptop with 12 GB of RAM of Intel(R) Core(TM) i7-5600U CPU @ 2.60GHz 2.59 GHz.

The algorithms were run 15 times and the precision, recall and f1-score results obtained in the best individual run are presented in Table \ref{tab:tab4}. Accordingly, the best learning rate according to the decision classes of the learning rates performed with different optimization techniques was realized in the PSO algorithm. Looking at the results of all algorithms individually, the "DOWN" class generally has a better learning percentage than the "UP" class. In the PSO algorithm, it is seen that the learning rates are smaller for both classes than in other algorithms.

\begin{table}[]
\centering
\caption{Classification performance results of the algorithms}
\label{tab:tab4}
\resizebox{\columnwidth}{!}{%
\begin{tabular}{|p{0.2\textwidth}|p{0.2\textwidth}|p{0.2\textwidth}|p{0.2\textwidth}|p{0.2\textwidth}|}
\hline
Algorithm                 & Classes & Precision & Recall & F1-score \\ \hline
\multirow{2}{*}{Grid}     & DOWN    & 0.87      & 0.84   & 0.85     \\ \cline{2-5} 
                          & UP      & 0.79      & 0.82   & 0.81     \\ \hline
\multirow{2}{*}{Random}   & DOWN    & 0.89      & 0.84   & 0.86     \\ \cline{2-5} 
                          & UP      & 0.79      & 0.85   & 0.82     \\ \hline
\multirow{2}{*}{Bayesian} & DOWN    & 0.87      & 0.89   & 0.88     \\ \cline{2-5} 
                          & UP      & 0.84      & 0.83   & 0.83     \\ \hline
\multirow{2}{*}{Genetic}  & DOWN    & 0.87      & 0.89   & 0.88     \\ \cline{2-5} 
                          & UP      & 0.84      & 0.82   & 0.83     \\ \hline
\multirow{2}{*}{PSO}      & DOWN    & 0.94      & 0.94   & 0.94     \\ \cline{2-5} 
                          & UP      & 0.92      & 0.91   & 0.92     \\ \hline
\end{tabular}%
}
\end{table}

The PSO algorithm has the best average test and training accuracy, as shown in Table \ref{tab:tab5}. The PSO performed very close to 1 in training accuracy.  GA and Bayes optimization algorithms presented similar results. Grid search and random search have worse performances as expected. Another criterion observed in this study is the CPU times of the algorithms. In some cases, search algorithms may require long periods of CPU-times, or it is not possible to perform these studies without highly capable processing power. The environment in this study could be completed by the days of these studies. Accordingly, when the CPU times are compared, it is seen that the CPU time in the Grid search is low because the search space in the Grid search consists of integer numbers. In addition, GA has been the algorithm that takes the most CPU time. PSO was concluded as the best performing algorithm in a reasonable time.

\begin{table}[hbt!]
\centering
\caption{Accuracy performances of optimization algorithms}
\label{tab:tab5}
\resizebox{\columnwidth}{!}{%
\begin{tabular}{|p{0.25\linewidth}|p{0.25\linewidth}|p{0.25\linewidth}|p{0.25\linewidth}|}
\hline
\textbf{Algorithm} & \textbf{Test Accuracy} & \textbf{Train Accuracy} & \textbf{CPU Time}     \\ \hline
Bayes     & 0.8611        & 0.8825         & 5850.78 \\ \hline
GA        & 0.8609        & 0.8822         & 6639.02 \\ \hline
Grid      & 0.8297        & 0.8409         & 714.89  \\ \hline
PSO       & 0.9262        & 0.9933         & 1286.4  \\ \hline
Random    & 0.8425        & 0.8586         & 2567.56 \\ \hline
\end{tabular}%
}
\end{table}

\section{Conclusion}

HPO has become necessary because of the high number of hyper parameters in machine learning applications. It is not possible to manually adjust the hyper parameters in machine learning. This only applies to problems with very small data sets or a very small number of parameters and solution spaces. Even in a mid-level machine-learning problem, the topic of HPO is complex and consists of many stages. Therefore, there is a need to develop, test and compare optimization algorithms to solve the HPO problem. Optimization algorithms can be classified in several ways. One of these classes is meta-heuristic algorithms. HPO studies powered by meta-heuristic algorithms may have higher performance, reliable and provide results without scanning the entire search space. Therefore, meta-heuristic algorithms offer an important application area for HPO. Random search, grid search, Bayesian optimization, genetic algorithms and PSO algorithms were tried on a generic data set discussed in this study. As a result of the study, it was shown that PSO algorithms are superior to other algorithms both in terms of learning performance and faster operation. In the studies to be carried out after this study, studies on which algorithm from meta-heuristic algorithms will have a performance in HPO studies with which search space will be investigated. In addition, various studies can be carried out by testing meta-heuristic algorithms, which are relatively new.

\bibliographystyle{unsrt}  
\bibliography{references}  
\end{document}